\documentclass[]{fairmeta}

\usepackage{amssymb}
\usepackage{xcolor}
\usepackage{tcolorbox}
\usepackage[utf8]{inputenc}
\usepackage[T1]{fontenc}
\usepackage{lmodern}
\usepackage{wrapfig}
\usepackage{tcolorbox}
\usepackage{array}

\usepackage{array}
\usepackage{makecell}
\usepackage{soul}

\newtcolorbox{questionbox}[1]{
    colback=blue!5!white,
    colframe=blue!50!black,
    fonttitle=\bfseries,
    title=#1
}

\newenvironment{significance}
{\begin{tcolorbox}[colback=gray!10,colframe=black!40,title=Forewords]}
{\end{tcolorbox}}

\title{\LARGE Why AI systems don't learn and what to do about it\\
\Large Lessons on autonomous learning from cognitive science}

\author[1,2]{Emmanuel Dupoux}
\author[3]{Yann LeCun}
\author[1,4]{Jitendra Malik}

\affiliation[1]{FAIR at META}
\affiliation[2]{École des Hautes Études en Sciences Sociales}
\affiliation[3]{NYU}
\affiliation[4]{UC Berkeley}

\abstract{
We critically examine the limitations of current AI models in achieving autonomous learning and propose a learning architecture inspired by human and animal cognition. The proposed framework integrates learning from observation (System A) and learning from active behavior (System B) while flexibly switching between these learning modes as a function of internally generated meta-control signals (System M). We discuss how this could be built by taking inspiration on how organisms adapt to real-world, dynamic environments across evolutionary and developmental timescales.}
\date{\today}
\correspondence{dpx at \email{@meta.com}}

\begin{document}
\maketitle

\begin{wrapfigure}{r}{0.45\textwidth}
\vspace{-12pt}
\begin{significance}
The dominant AI research paradigm today relies on hyperscaling of text-based LLMs with ever larger models, data and compute. But even prominent architects of this approach such as Ilya Sutskever\footnote{\href{https://www.businessinsider.com/openai-cofounder-ilya-sutskever-scaling-ai-age-of-research-dwarkesh-2025-11?utm_source=chatgpt.com}{Business insider}} and Andrei Karpathy \footnote{\href{https://navaneethsen.medium.com/the-andrej-karpathy-interview-with-dwarkesh-patel-c10659db456c}{Medium}} suggest we may be hitting diminishing returns.  Areas of concern include (1) confronting the "data wall" on quality text data (2) inability to learn new things beyond current human knowledge because of the absence of interaction with the environment \citep{silver2025welcome} (3) excessively language-centrism as opposed to spatial, embodied and grounded reasoning in the physical world (4) lack of continual life-long learning (self-improvement after deployment). While these critiques echo long standing controversies within cognitive science on the non-verbal cognition \citep{johnson1983mental}, and situated interactions \citep{piaget1952origins,vygotsky1978mind} in intelligence, it behooves us as scientists to take stock of progress from both fields and look beyond the current paradigm. What could come next?
\end{significance}
\vspace{-35pt}
\end{wrapfigure}

Both AI and Cognitive Science emerged in the 1950's in the post-war intellectual ferment which brought together neural modeling, computation, information, and control. While the objectives  differ --creating intelligent machines vs. scientific understanding of brains and behavior, the intellectual trajectories of AI and cognitive science have overlapped with varying degree of cross-fertilization. Today, the successes of Deep Learning ushered an era of deeper cross-disciplinary interactions. As AI models tackle high level human abilities like language, visual understanding and reasoning, they incorporate concepts and evaluation methods from the cognitive and neural sciences. Conversely, AI systems provide sorely needed quantitative theories of cognitive processes that can be tested against empirical data. {\em Paradoxically, given the importance of deep learning, one key component of human intelligence remains out of reach for current AI models: the ability to learn as humans do.}

\vspace{-0.em}
\section{What is autonomous learning?}
\vspace{-0.em}

Consider the distinction between children and current AI models. Children learn and act from birth. They flexibly choose what to attend, what to learn, when to act or observe, and more generally how to switch between different learning modes \citep{Botvinick2019HierarchicalRL,Shenhav2017ExpectedValueOfControl}. For example, a toddler trying a new toy may explore it randomly (\textit{learning through action}; \citealt{Gopnik2017}), or by watching a peer, attempt to imitate the goal or gesture, depending on context (\textit{learning through observation}; \citealt{Gergely2002RationalImitation,Tomasello1999CulturalLearning}). They may follow a caretaker's verbal instruction on how to use the toy (\textit{learning through communication}; \citealt{Csibra2009NaturalPedagogy}), or take a pause and daydream about the various ways to use the toy (\textit{learning through imagination}; \citealt{Redshaw2016FutureThinking}). 

In contrast, AI models, once deployed, learn essentially \textit{nothing}; their mode of operation is fixed, and if not adapted to their environment, a new model has to be rebuilt using new data by human experts-in-the-loop \citep{Hadsell2020Continual}.
Furthermore, the different learning modes exemplified in children are typically siloed into distinct machine learning paradigms (e.g., self-supervised learning, supervised learning, reinforcement learning), each requiring specific data curation pipelines and training recipes; when the different modes are mixed, it is mainly through rigid sequences of training recipes established through trial and error by human experts and tuned to particular applications (chatbots, coding assistants, etc.). 
In other words, in current AI systems, learning is \textit{outsourced} to human experts instead of being an \textit{intrinsic} capability.

Arguably, the inability to learn may explain some difficulties of AI systems to be deployed in real life. AI systems are built by optimizing an objective over a fixed set of \textit{training data}, typically lifted from the internet. However, once deployed in real life, this system may be confronted with new data that diverge significantly from this distribution, with unpredictable consequences. 
This phenomenon known as \textit{domain mismatch} cannot be fixed by merely increasing the training set size, as real-life data always contains new, unseen cases (\textit{heavy tailed}) and keeps changing over time (\textit{non-stationarity}) \citep{Geirhos2020Shortcut,Koh2021WILDS}. 
Modern AI addresses domain mismatch by breaking down model training into two phases: \textit{pretraining} using large generic datasets, and \textit{fine tuning} using data more appropriate to the target application \citep{Bommasani2021Foundation}. This is a first step in the right direction, but it still requires considerable human involvement, and there is no guarantee that this will work, as the system has not been primarily built to be fine-tunable or adaptable, especially on unfiltered raw data. 
In contrast, in biological organism, domain mismatch is mitigated by enabling the agent to learn and adapt from the data directly available in its environment,  allowing for species-specific cognitive adaptability. Humans are particularly adaptable in this respect, having shown the ability to spread very quickly to a variety of different ecological niches \citep{Boyd2011CulturalNiche}.

\begin{figure}
    \centering
    \includegraphics[width=0.72\linewidth]{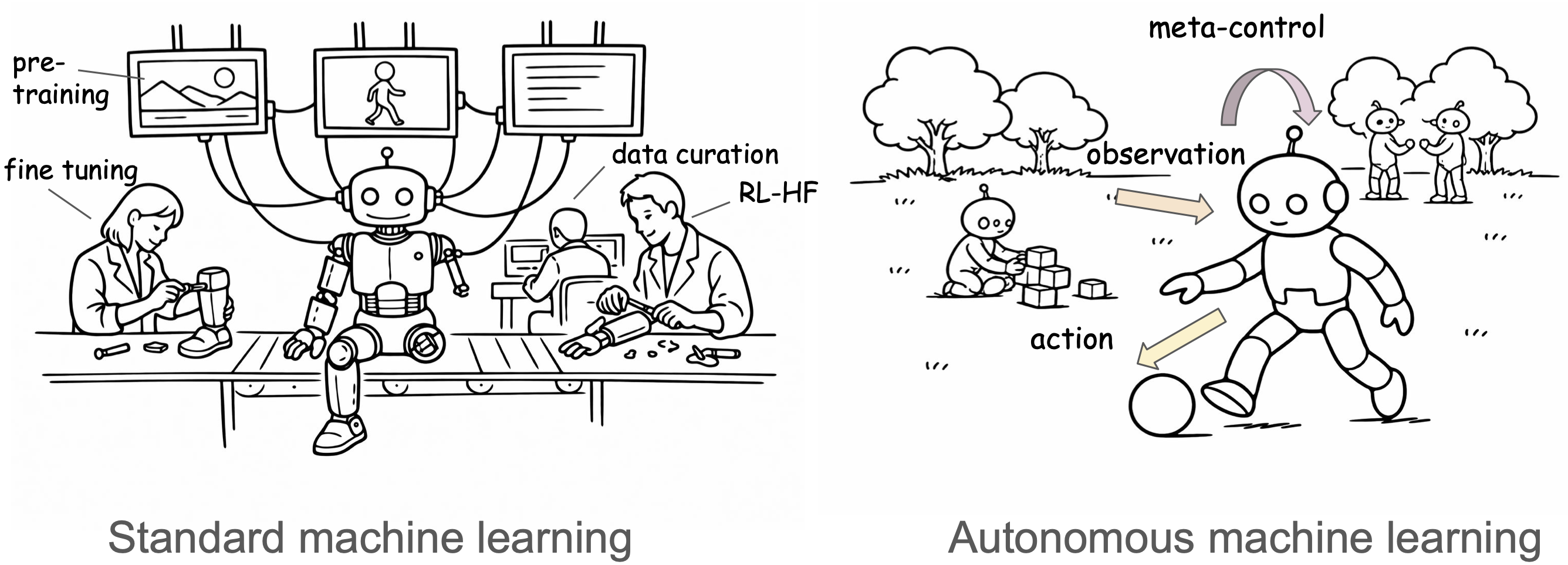}
    \raisebox{0.5cm}{\includegraphics[width=0.22\linewidth]{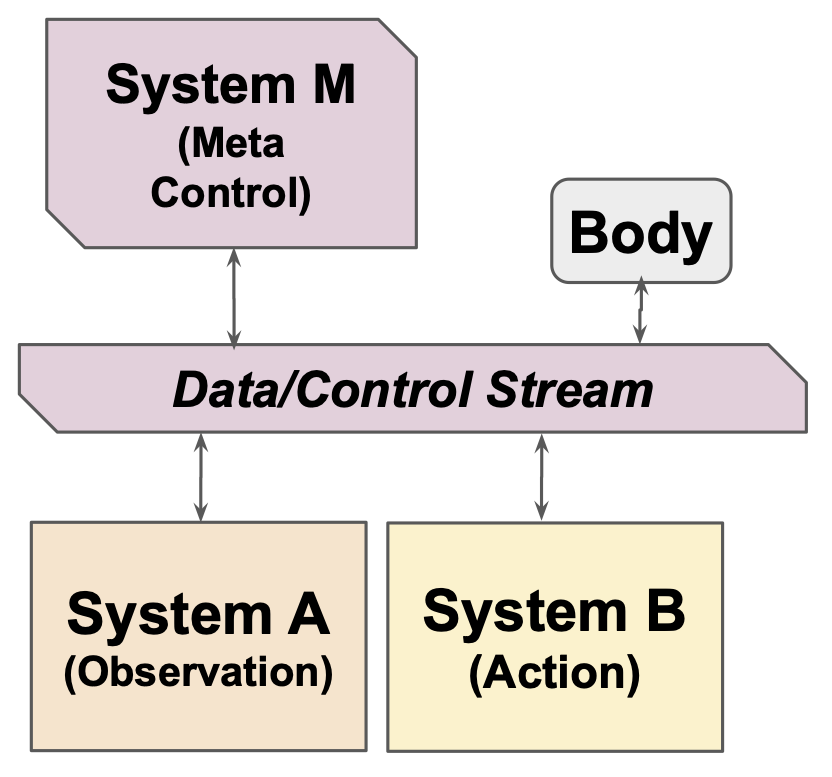}}
    \caption{\textbf{Standard machine learning (left)}. The machine does not learn by itself; it requires an assembly line of research engineers and data scientists collecting, formatting, and curating different kinds of data, each used to train successively different components of the model, each with specificly engineered loss and reward functions. The machine is then left with no ability to learn from its experience. \textbf{Autonomous machine learning (right).} The agent is learning directly in interaction with the world; the sources of data are generarated by the agent itself through different learning modes (learning by observation, by action, which can be extended to higher modes like learning by verbal interaction or self-play). Our proposed architecture include a meta controler enabling learning while operating in the real world. \textit{(Drawings from  \textit{ChatGPT})}.}
    \label{fig:placeholder}
\end{figure}

In this paper, we start from the idea that
autonomous learning should be considered a \textit{core capability}, essential for building reliable AI systems that can operate in the real world. Conversely, we take it that the development of adaptable AI systems can benefit cognitive science by providing quantitative models capable of addressing long-standing debates about the nature and origins of human intelligence. The main contributions of the paper are the following: we identify three conceptual and technical roadblocks that have so far limited the development of autonomous learning, and we propose possible diractions to address them. This should be as a high level roadmap which we hope will be useful in inspiring future cross-disciplinary work and collaborations.  

The first roadblock is conceptual: existing approaches to learning remain \textit{fragmented} across subfields, making it difficult to integrate them within a unified framework. In Section  \ref{sec:AB}, we argue that a path for integration is first to recognize two fundamental learning modes: learning through observation (\textit{System A}) and learning through action (\textit{System B}), and then classify the different ways in which these modes can interact with one another.  The second roadblock is the \textit{externalisation} of learning which is currently practiced in AI. To address this, we propose in Section \ref{sec:M} a \textit{meta-control architecture (System M)} that coordinates information flow between the learning components enabling to reproduce automatically the learning and data filtering recipes typically done by hand. We show that such system open up the possibility of higher order learning modes only found in some large brain species, like learning by through communication and imagination.  The third roadblock is the lack of effective methods to build such architectures at scale. In Section \ref{sec:evodevo}, we propose an \textit{evolutionary} inspired \textit{bilevel optimization} approach to jointly learn the meta-control model and the initial states of the system A and B components to achieve robust real-world behavior. In Section \ref{sec:conc}, we conclude by reviewing recent progress and outlining promising directions for research at the interface of AI and cognitive science in this emerging area of autonomous learning.

\section{Integrating Observation and Action}\label{sec:AB}

The ability to learn has been studied within a rich set of traditions and methods in the natural, social, and formal sciences. For lack of space, we will not attempt even a cursory review of this vast topic here, but rather point out one major fault line across these fields that separates what can be called \textit{observation-based} versus \textit{action-based} learning. In the first case, the organism is conceived passively accumulating sensory input, and learning through building a statistical or a predictive model of its input data \citep{Saffran1996Statistical,Rao1999Predictive}. We call the set of learning mechanisms that fall in this bucket: \textit{System A}. In the second case, the organism is conceived as an agent interacting in the world and trying to achieve a particular goal through the adjustment of its actions against the observed feedback from the environment \citep{Sutton2018,schultz1997neural}. We call the relevant set of mechanisms \textit{System B}\footnote{This terminology is reminescent of Kahneman's System 1 and 2, but is actually orthogonal to it, see Appendix \ref{sec:system12} for a discussion.}.

For methodological or historical reasons, these two conceptions have yielded distinct subfields within each of the relevant sciences, with little interaction between them (even using different terminology, as in Table \ref{tab:sysA} for System A). In the following sections, we describe their strengths and limitations and then outline why they need to be combined to account for how learning really takes place in living organisms.

\subsection{System A: Learning from Observation}

From a cognitive point of view, System A learning situations abound in early childhood where infants' abilities to act on the world are limited. 
For instance, infants initially can discriminate faces from multiple species at 6 months (e.g., human and monkey faces), but by 9 months they become specialized for human faces and lose sensitivity to non-human faces \citep{pascalis2002face}. Similarly, while newborns can distinguish phonetic contrasts from many languages,  between 6 and 12 months their perception improves for the sounds of their native language and degrades for non-native ones \citep{werker1984cross}. Infants are thought to learn a model of the world through observation, making them able to predict the future based on past observations: they are surprised by 'impossible' events or magic tricks \citep{spelke1992origins,baillargeon2004infants,spelke2007core}. All of these phenomena have been attributed to learning mechanisms capable of extracting the statistical distribution of sensory stimuli and/or predicting future stimuli conditioned on past events, both being instances of, both being instance of System A algorithms \citep{saffran2018infant}. 

\begin{table}[h]
\caption{Sample observational learning problems (System A) from different sensory inputs, proposed mechanisms with corresponding terminology in Cognitive Science and AI fields and sample AI algorithms (non-exhaustive).}\label{tab:sysA}
\begin{tabular}{>{\raggedright\arraybackslash}p{1.2cm} >{\raggedright\arraybackslash}p{2.8cm} >{\raggedright\arraybackslash}p{3.2cm} >{\raggedright\arraybackslash}p{3.2cm} >{\raggedright\arraybackslash}p{4cm}}

\toprule
& & \multicolumn{2}{c}{\bf Proposed Learning Mechanisms}&\\
\cline{3-4}
\bf Input & \multicolumn{1}{l}{\bf Task} & \multicolumn{1}{l}{\bf CogSci terminology} & \multicolumn{1}{l}{\bf AI terminology}&\bf (sample) AI Algorithms \\
\toprule
Speech   & Learning phonetic categories  & Distributional Learning /\newline Perceptual Learning   &  Self-Supervised Learning / Acoustic Unit Discovery & CPC \citep{Oord2018}, HuBERT \citep{Hsu2021HuBERT}\\
\hline
Language & Syntax acquisition & Statistical Learning  & Language Modeling & GPT {\citep{Radford2018GPT}}, BERT {\citep{Devlin2019}}, GSLM {\citep{Lakhotia2021GSLM}} \\
\hline
Images   & Face recognition, Object categorization  & Perceptual Learning & Self-Supervised Learning & SimCLR {\citep{Chen2020}}, MoCo {\citep{He2020}}, DINO {\citep{Caron2021}}, I-JEPA {\citep{Assran2023IJEPA}} \\
\hline
Video    & Intuitive physics & Perceptual Learning  & Predictive World Modeling      & PredNet \citep{Lotter2017PredNet}, V-JEPA \citep{bardes2024revisiting}\\
\hline
Language \newline + Vision & Learning words and sentence meaning & Cross-situational Learning & Multimodal Language Modeling  & CLIP {\citep{Radford2021}}, Flamingo {\citep{Alayrac2022Flamingo}},  Transfusion {\citep{Zhou2024Transfusion}} \\              
\bottomrule   
\end{tabular}
\end{table}

From an AI perspective, such algorithms include Self-Supervised Learning (SSL) models applied to static datasets or passively collected sensory streams \citep{Chen2020,He2020,Devlin2019}. These paradigms can be classified based on their modality, data type, and structure. Some systems operate on a single modality such as text, images, or audio, while others combine modalities, for instance, vision and language \citep{Radford2021}. Some work on symbolic, discrete data like tokens, while others learn directly from continuous sensory input. Finally, some models explicitly exploit the spatial or sequential structure of their inputs, such as grids or time series.

Abstractly, one can describe this class of algorithms in the following way. 
Let data come from a distribution  $\mathcal{D}$ ($x \sim \mathcal{D}$). We define a \textit{task generator} $\mathcal{G}$
that, given a raw sample $x$, produces an input–target pair:

\begin{equation}
(x_{\text{in}}, x_{\text{tar}}) = \mathcal{G}(x)
\label{eq:G}
\end{equation}

We want to learn a representation
$z=f_{\theta}(x)$ with parameters $\theta$ that minimizes a loss. The training objective is:

\begin{equation}
\theta^\star = \arg\min_\theta \; \mathbb{E}_{x \sim \mathcal{D}} \; \mathcal{L}\big(f_\theta(x_{\text{in}}), x_{\text{tar}}\big)
\label{eq:A}
\end{equation}

System A has several strengths. It scales well with large datasets and is capable of discovering abstract latent representations that can be organized hierarchically across different levels of abstraction, from low-level sensory features to high-level conceptual categories \citep{Bengio2013}, and can support robust transfer to downstream tasks \citep{Devlin2019,Caron2021}.

However, it also faces limitations, as is apparent in the formulations in Equations \ref{eq:G} and \ref{eq:A}. System A models need access to $\mathcal{D}$ and a task generator $\mathcal{G}$, both of which typically require considerable human expertise, and have to be tailored to each domain with care \citep{Tian2020GoodViews}. They lack any built-in mechanism to decide what data might be useful or should be acquired next \citep{Settles2009ActiveLearning}. Furthermore, their representations are disconnected from the agent's ability to act, making it hard to ground what they learn in real-world behavior \citep{Bisk2020ExperienceGrounds}. Last but not least, because they are based purely on observation, they struggle to distinguish between correlation and causation \citep{Scholkopf2019Causality}.

\subsection{System B: Learning from Action}


From a cognitive point of view, typical System B learning situations can be found in basic motor learning in children, like learning to walk. Here, observation of the world does not necessarily help, as children do not initially attempt to imitate other agents' mode of locomotion \citep{Adolph2012LearningToWalk}. Rather, through trial and error they go through various stages using non-bipedal modes (rolling, crawling), before they develop the ability to stand and take a few wobbling steps and finally develop a mature gait. To a certain extent, vocal learning follows a similar path, whereby initial vocal explorations by infants are not similar to their linguistic input, and even arise in children with hearing loss \citep{Oller1988Babbling}.

From a machine learning point of view, the class of System B algorithms comprises learning mechanisms that operate through interaction. Acting is to intervene on the world through a sequence of \textit{actions} $a_t$, to reach a given \textit{goal} (optimizing some \textit{reward} $r$ over some \textit{time horizon} $T$). The world is characterized in terms of its \textit{states} $s_t$, and \textit{transition dynamics} $M(s_{t+1}|s_{t},a_{t})$. The transition dynamics of the agent is called a \textit{policy} ($\pi$). If the world dynamics were totally known in advance and relatively simple, the optimal sequence of actions could be derived mathematically without any learning all (as in \textit{control theory}, \citealt{Bertsekas2019RLControl}). If the world dynamics are unknown, then the system has to learn about it to optimize its actions (as for \textit{reinforcement learning} and \textit{planning}; \citealt{Sutton2018,Russell2020AI,Moerland2023MBRL}). The general problem can be formulated:

\begin{equation}
\operatorname{Maximize} J(\pi) =  
\mathbb{E}_{} 
\left[ \sum_{t=0}^{T}\gamma^{t} r(s_{t},a_{t}) \right]
\end{equation}

Where:

\begin{table}[h!]
\center
\begin{tabular}{lp{6cm}}
 $s_{t+1} \sim M(.| s_{t}, a_{t})$  & Dynamic model of the world\\
 $a_{t} \sim \pi(.| s_{t})$  & Agent's policy\\
 
 $r(s,a)$& Reward function\\
 $\gamma \in [0,1)$& Discount factor\\
\end{tabular}
\end{table}

\begin{table}[h]
\caption{Popular paradigms addressing System B optimization problems, differing in how the world model $M$ and the policy $\pi$ are defined, whether they are learned from data in a training phase, and how they are used at inference time, with selected examples of applications in the modeling of animal behavior.}\label{tab:sysB}

\begin{tabular}{>{\raggedright\arraybackslash}p{1.2cm} >{\raggedright\arraybackslash}p{2.8cm} >{\raggedright\arraybackslash}p{3.2cm} >{\raggedright\arraybackslash}p{3.2cm} >{\raggedright\arraybackslash}p{4cm}}

\toprule
\bf {Paradigm}    &\bf {W. Model $M$\newline (train., infer.)}  & \bf{Policy $\pi$\newline (train., infer.)}    & \bf {When is optimiza-\newline tion solved?}  &\bf {Example}   \\

\toprule

{\bf Control\newline \bf Theory}    & {fixed\newline (no,no)}     & {derived analytically\newline (no, direct)}& offline (design time)  & {Spinal reflexes, \newline saccadic eye movement\newline {\citep{Robinson1981Oculomotor}}}\\
\hline
{\bf Adaptive\newline \bf Control}  & {adaptive param.\newline (yes, no)}     & {derived analytically\newline (no, direct)} & {online (parameter\newline  estimation)} & {Motor adaptation \newline {\citep{Shadmehr1994Adaptive}}}\\
\hline
{\bf Model-Free\newline \bf RL}    & none                               & {NN\newline (yes, direct)} & {training time (compil-\newline ation of experience \newline into reactive policy)} & \{Habitual actions \newline{\citep{schultz1997neural}}\\
\hline
{\bf Model-Based\newline \bf RL}    & {NN (yes,\newline no / unrolled )}& {NN (yes, \newline direct / search)} & {both training \newline and inference} & {Goal-directed behaviors,\newline foraging strategies \newline {\citep{Daw2011ModelBased}}} \\   
\hline
{\bf Planning}          & {simulator / NN\newline (no/yes, unrolled)}   & {computed online\newline (no,search)}   & {inference time \newline (thinking before acting)} & {Detour planning, \newline mental simulation\newline {\citep{Pfeiffer2013Hippocampal}}} \\                    
\bottomrule
\end{tabular}
\end{table}

Methods to optimize the objective function depend on whether the world transition dynamics and reward function are known or must be learned, and whether one searches through a space of actions, or learns a policy that directly predicts the next action to be taken given the world state (see Table \ref{tab:sysB}). These systems also vary in their reward sources, which may be provided externally through task-specific signals or generated internally through curiosity, novelty, or empowerment \citep{Schmidhuber1991,Pathak2017Curiosity,Mohamed2015}. The size and structure of the action space also vary widely. Simple environments use a small set of discrete actions, but real-world tasks often require complex, continuous, and high-dimensional action spaces \citep{Lillicrap2015}. Exploration strategies can be random, curiosity-driven, or guided by a goal or policy \citep{Ecoffet2021}.

System B has important strengths: it is grounded in control and interaction, enabling it to learn directly from sparse or delayed outcomes, making it naturally suited for real-time and adaptive behavior. It can also discover truly novel solutions via search \citep{Silver2017AlphaGoZero}. However, it also faces major limitations. Primarily, it is notoriously sample-inefficient, often requiring large numbers of interactions to learn even simple tasks \citep{DulacArnold2021RealWorldRL}. It struggles in high-dimensional or open-ended action spaces. Furthermore, it depends on having well-specified reward functions and interpretable actions, which are rarely available in naturalistic settings \citep{Amodei2016ConcreteSafety}.

\subsection{System A Helping System B}

At an intuitive level, learning through action is easy when the number of possible actions is limited, and the world states are easy to track. This is typically the case in games like chess or video games. This is less the case in real life where the action space grows exponentially with the number of degrees of freedom (roughly 200 to 300 for robotics or animation), and world states are virtually unlimited. In humans and animals, one dominant idea is that the sensory/motor system provides a kind of curriculum by limiting the resolution of sensors at birth (children are myopic) and the effective degrees of freedom (with very synergistic muscles) \citep{TurkewitzKenny1982,Bernstein1967Coordination}. Even there, the search space is vastly larger than in a game of chess, and this is where System A can help by providing compressed representations for states and actions, predictive world models, and intrinsic reward signals that would make learning and planning more tractable \citep{Ha2018WorldModels,Yarats2021CURL}.

\textbf{Abstract representations of states and actions}. By observing the world as experienced by the agent, System A can learn abstract representations of observations through SSL methods. This can be used as a proxy for the representation of \textit{world states} that are more abstract and more compact than raw sensory data (pixels or sound waves). For instance, CURL \citep{laskin2020curl} uses contrastive vision pretraining to derive a compact representation from pixels that can be fed to an RL agent learning policies for Atari games, with  performance equivalent to an agent trained on hand-coded world states. (see also ACT; \citealt{Zhao2023ACT}). Similarly, instead of working from raw pixels, many robotics papers leverage a pretrained vision encoder to provide useful features \citep{Nair2022R3M,Radosavovic2022MVP}.

Similarly, by observing sequences of actions taken by the agent, SSL techniques can yield abstract representations of the \textit{action space} or group successive actions into skills (see DIAYN, \citealt{eysenbach2019diversity}; action chunking, \citealt{li2025reinforcement}). These principles are applied to robotics to reduce the dimensionality of action space  (eg. CLAM, \citealt{liang2025clam}). 
In \cite{radosavovic2024humanoid,radosavovic2024learning},  generatively modeling sensorimotor trajectories
in humanoid locomotion (in simulation), facilitates subsequent RL and application to real ,challenging terrain.

Forward looking, System A can learn latent action spaces from unlabelled video (eg. LAWM; \citealt{tharwat2025latent}) or infer \textit{goals} or \textit{rewards} from raw videos, enabling inverse RL and direct imitation of tasks or goals \citep{sermanet2016unsupervised,ma2023vip}.

\textbf{Predictive World Models}. Among System A models,  predictive models learn to predict future states based on past states, capturing the dynamics of the environment. This addresses a critical problem in (model-free) RL, which is the combinatorial explosion of the search space. Predictive models, when conditionned on self-actions can turn System B into model-based RL, enabling planning instead of blind trial-and-error. 

Notable models include PlaNet \citep{Hafner2019planet}, Dreamer \citep{Hafner2020dreamer} and SPR \citep{schwarzer2020data}. The ability to learn an internal simulation that can fully replace an external environment has been demonstrated in video games and Go with Mu-zero \cite{schrittwieser2020mastering}. The scaling of these ideas to more complex environment is still underway. Predictive World models are split between pixel-based generative models (e.g., Genie \citep{bruce2024genie}, Gato \citep{reed2022generalist}), and Video Joint Embedding Predictive Architecture (V-JEPA) that predict in latent space \citep{bardes2024revisiting}, enabling better modeling of physics \citep{garrido2025intuitive}, and showing fast transfer to robotics applications \citep{assran2025vjepa2selfsupervisedvideo}.

\textbf{Intrinsic reward signals}. Reinforcement Learning hasclassically been confronted with the exploration/exploitation dilemma, where exploitation tries to optimize immediate reward, but may fail because of incomplete knowledge about the effect of actions on the world, and exploration improves world knowledge but may delay immediate reward. System A can help by providing intrinsic reward signals like prediction errors, uncertainty, or novelty, enabling the agent to explore efficiently, and shift to exploitation once it is confident \citep{oudeyer2007intrinsic,schmidhuber2010formal,aubret2023information}. These ideas have been applied to the video game domains (e.g., \citealt{Pathak2017Curiosity,badia2020never,kayal2025impact}), but applications to robotics remain limited \citep{tang2025deep,taylor2021active}.

Each of these mechanisms reduces the burden on System B by simplifying the search space of model-free RL, by reducing the dimensionality of representations, providing forward models to guide search, and exploration rewards to reduce uncertainty. Much remains to be done in scaling these different approaches to real-life situations.

\begin{figure}[t]
    \centering
    \includegraphics[width=10cm]{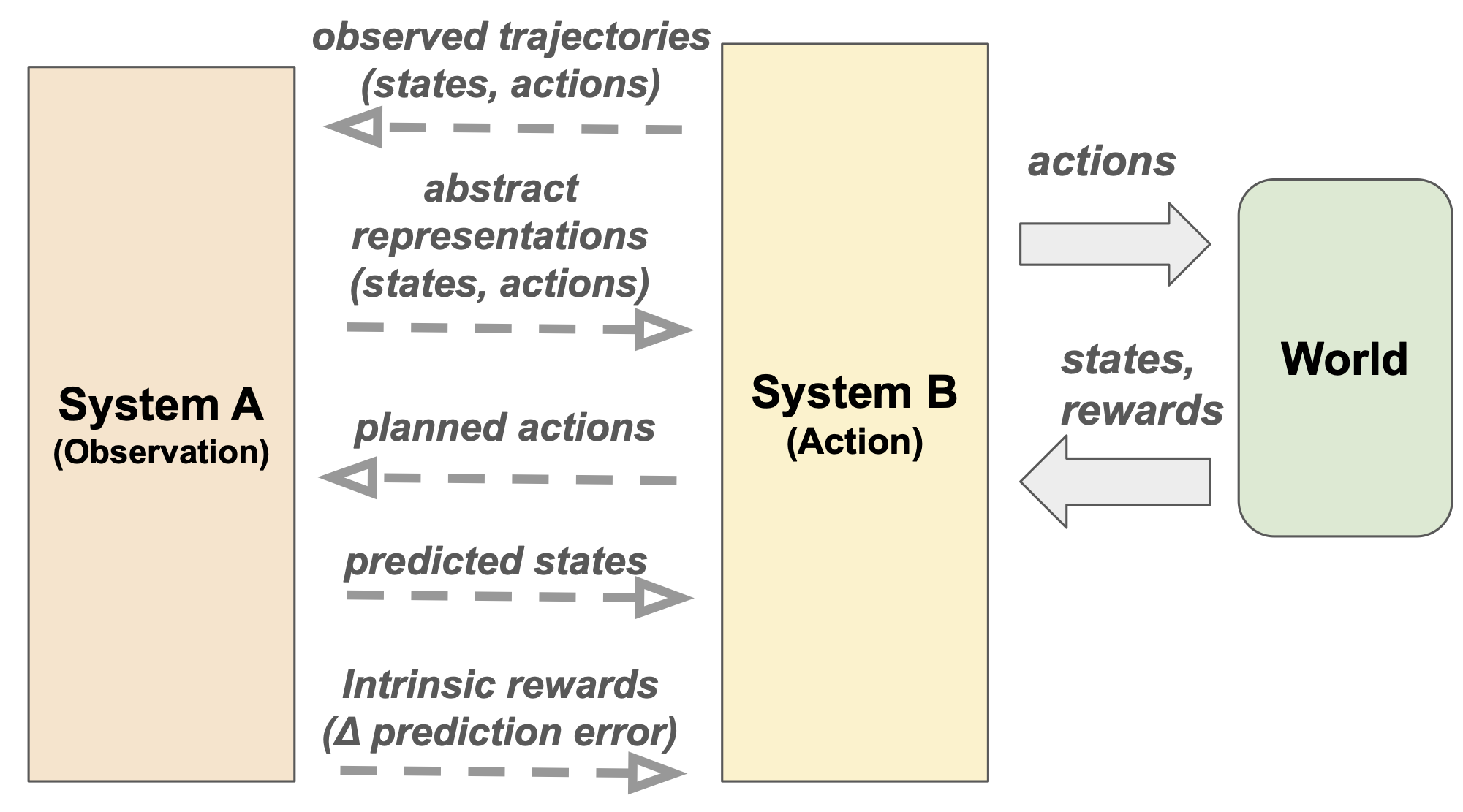}
    \caption{Summary of modes of interactions between Systems A and B. System A provides System B with predictions of future states conditioned on past states and actions, with hierarchical abstractions over possible actions, and a SSL loss that can be used for curiosity/exploration. System B through its action provides rich and task relevant input for System A to learn from.}\label{fig:interactionsA&B}
\end{figure}

\subsection{System B Helping System A}

System A's primary limitation is its reliance on passive or static data. Without guidance or data curation, it may fail to learn useful representations from uninformative, noisy, or irrelevant data streams \citep{lavechin2023babyslm,oquab2023dinov2,gadre2023datacomp}. But even human newborns are not passive observers: they can affect their data stream by orienting their visual and auditory attention to specific parts of the sensory field (e.g., orientation towards faces or speech sounds; \citealt{morton1991,vouloumanos2007}). As they grow in motor autonomy, they are more and more able to seek out specific sources of information useful for their goals.
System B, through active behavior, can help collect better data and provide grounding for learned representations. \citet{gibson1966senses}'s notion of active perception: “We see in order to move and we move in order to see” is a classic statement of the active gathering of information driven by a goal.

System B can support System A learning in two fundamental ways: directly, by helping to optimize System A’s own predictive objectives (\textit{active SSL}), or indirectly, by exploring the environment in ways that yield task-relevant or informative trajectories (\textit{goal-directed SSL}).

\textbf{Active self-supervised learning}. System B is explicitly optimizing System A’s ability to represent or generalize. For instance, through eye or head movement, System B can select a particularly 'interesting' portion of the sensory data to learn from, 'interesting' being defined by System A itself such as uncertainty, prediction error, or learning progress \citep{Gottlieb2013Information,Oudeyer2007IntrinsicMotivation}. This resembles curriculum learning or active data selection, but achieved through actions in the world \citep{Smith2018Curriculum}. 
This logic can be extended to interventions that help disambiguate perception by revealing causal relationships that would be missed through passive observation \citep{Agrawal2016}.

\textbf{Goal-directed self-supervised learning}. System B optimizes its own task-related reward, and provides data to System A as a by-product. This results in data that may not be optimized for System A’s loss but nonetheless offers rich and grounded input that supports the development of task-relevant representations \citep{Pathak2017Curiosity,Pong2019}.

In both of these cases, System B may enrich the data available to System A by generating parallel action/perception datasets that support cross-modal learning or even a form of supervised learning, to the extent that actions are lower-variance and more reliable than noisy, ambiguous sensory inputs. 

\vspace{-0.7em}
\subsection{Towards deeper integration of learning modes}

\begin{figure}
    \centering
    \includegraphics[width=0.7\linewidth]{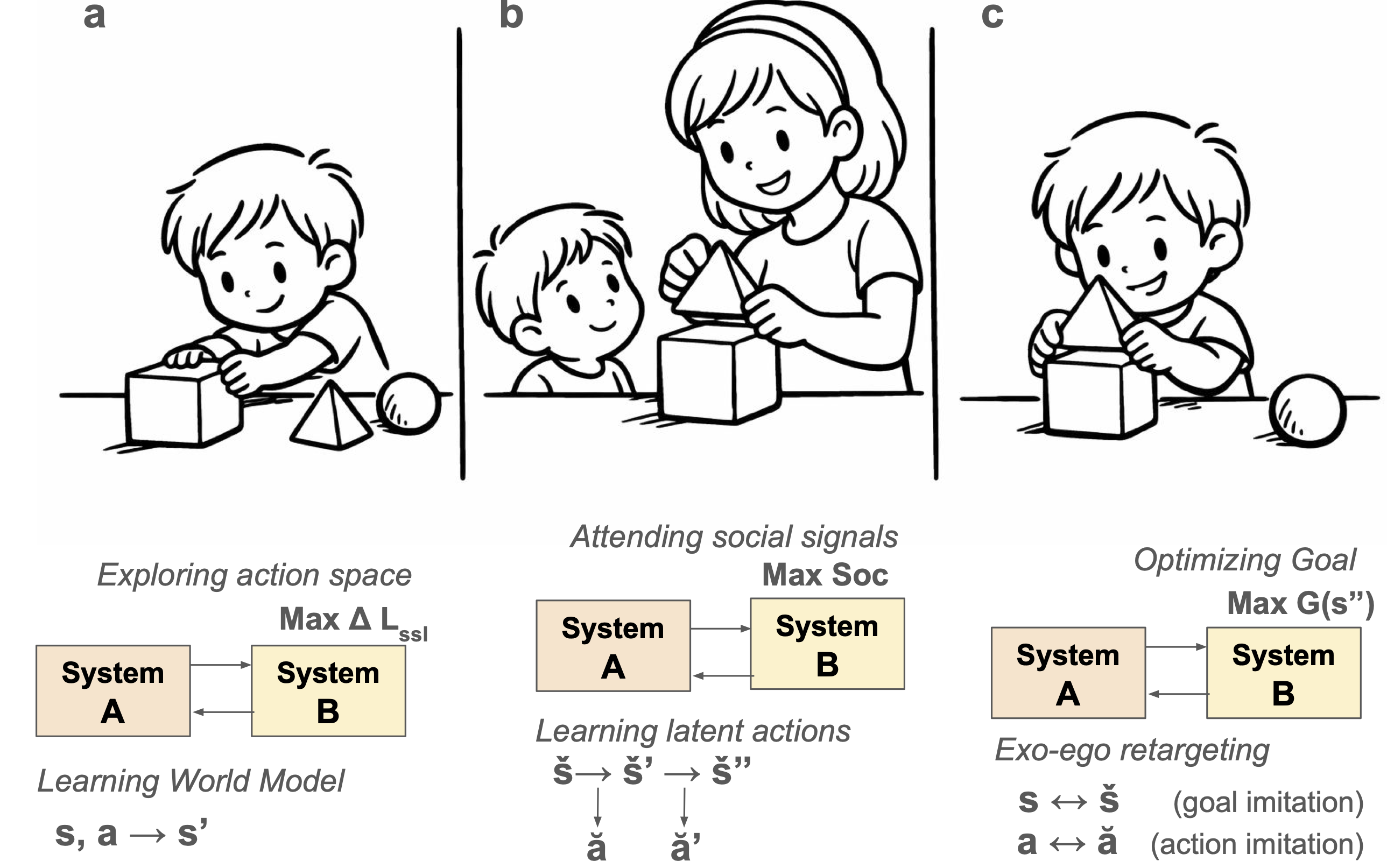}
    \caption{Interactions between learning modes for imitation learning. (a) \textbf{Self Play.} System B provides action, state trajectories to System A that learns a World Model, and provides a prediction-based intrinsic reward signal to system B. (b) \textbf{Social Observation}. System B directs the attention to peers that provide System A with complex trajectories from which it infers latent actions.  (c) \textbf{Retargeted imitation}. System A learns to map exocentric actions and states to egocentric ones, helping system B to achieve goal-directed behavior. \textit{(image from ChatGPT)}}
    \label{fig:imit}
\end{figure}

Figure \ref{fig:interactionsA&B} summarizes the interactions between the two systems discussed so far. In AI, deep integration between System A and System B has already been succesful in constrained domains. In games, agents like MuZero \citep{schrittwieser2020mastering} and Dreamer \citep{Hafner2023DreamerV3} couple learned latent dynamics with action planning to achieve superhuman performance. Such integration is gaining traction in robotics where vision-language-action models leverage massive, passively trained representations to directly guide motor execution \citep{Driess2023PaLME,Brohan2023RT2}.
However, in current systems, the learning recipe and runtime execution remains rigidly fixed by human engineers, while in living organisms, System A and B interact far more autonomously and fluidly. Classical AI has proposed \textit{cognitive architectures} (ACT, SOAR; \citealt{Anderson2004ACT,Laird2012SOAR}) to formalize such flexibility, which have recently been adapted to deep learning. \cite{Lecun2022Path} proposes an architecture integrating self supervived learning of world models and planning, within an energy based approach, enabling flexible specification of tasks through a \textit{configurator}. Flexbility is also the target of \textit{Global Workspace Theory} architectures \citep{Goyal2020Inductive}, which are still in early stage of developmennt \citep{Goyal2022Coordination}.

In this section, we exemplify the flexibility problem with \textit{imitation learning} (also called \textit{social learning})—a complex behavior requiring continuous toggling between passive observation and active motor correction—before turning to our proposed architecture, System M.

Imitation learning consists of reproducing an action performed by a conspecifics. It has been documented in several species \citep{whiten1992nature} and in human children \citep{meltzoff2007like,rizzolatti2004mirror}. Unpacking this capability reveals that it relies on tightly integrated Systems A and B learning modes (see Figure \ref{fig:imit}). The organism needs to learn a sequence of observed action in conspecifics (System A world modeling), then map this sequence to a corresponding sequence of its own actions. This is not obvious, because the space of obervations (images) is not the same as the space of actions (motor commands). As the bodies between learner and demonstrators are not the same, this creates a 'retargeting problem', which can be addressed by trying to reproduce the observed state changes (goal imitation) or by learning a correspondence between exocentric and egocentric actions (action imitation). Solving this problem involves learning a common world model based on self-play and social observation, and learning a policy to reproduce these actions.  These phases of observation and action likely alternating in  cases of complex skills requiring a hierarchy of subskills.

Current approaches in robotics also learn from human examples, but they sidestep the retargeting problem  by using data from \textit{teleoperation} (where humans directly provide motor actions for the agent \citealt{Fu2024MobileAloha}). This limits the scale at which new skills can be added to robots, as well as the speed and agility of the learned skills. Recent approaches pretrain models using video data of human actions and learn latent actions \citep{Baker2022VPT,tharwat2025latent}. This  reduces the amount of teleoperation data without eliminating it. A more flexibly integrated System A and System B model would be able to learn from videos of humans without any teleoperation data. In the next section, we explore what such a flexible architecture could look like.

\section{Meta-control for autonomous learning (System M)}\label{sec:M}

In humans and animals, System A and System B are active from birth, exhibiting a continuous, fluid, and autonomous interaction across cognitive domains (e.g., visuo-motor skills, social skills, and language). In current AI systems, the switching between learning modes is managed offline through rigid training recipes designed and controlled by human experts in a pipeline generally referred to as \textit{MLOps} \citep{zhengxin2023mlopsspanningmachinelearning}. A quintessential example is the training pipeline of Large Language Models (LLMs), which strictly forces a massive phase of unsupervised next-token prediction (System A) followed by a subsequent, disconnected phase of reinforcement learning from human feedback (System B; \citealt{ouyang2022training}).

\begin{figure}[t]
    \centering
    \includegraphics[width=0.3\linewidth]{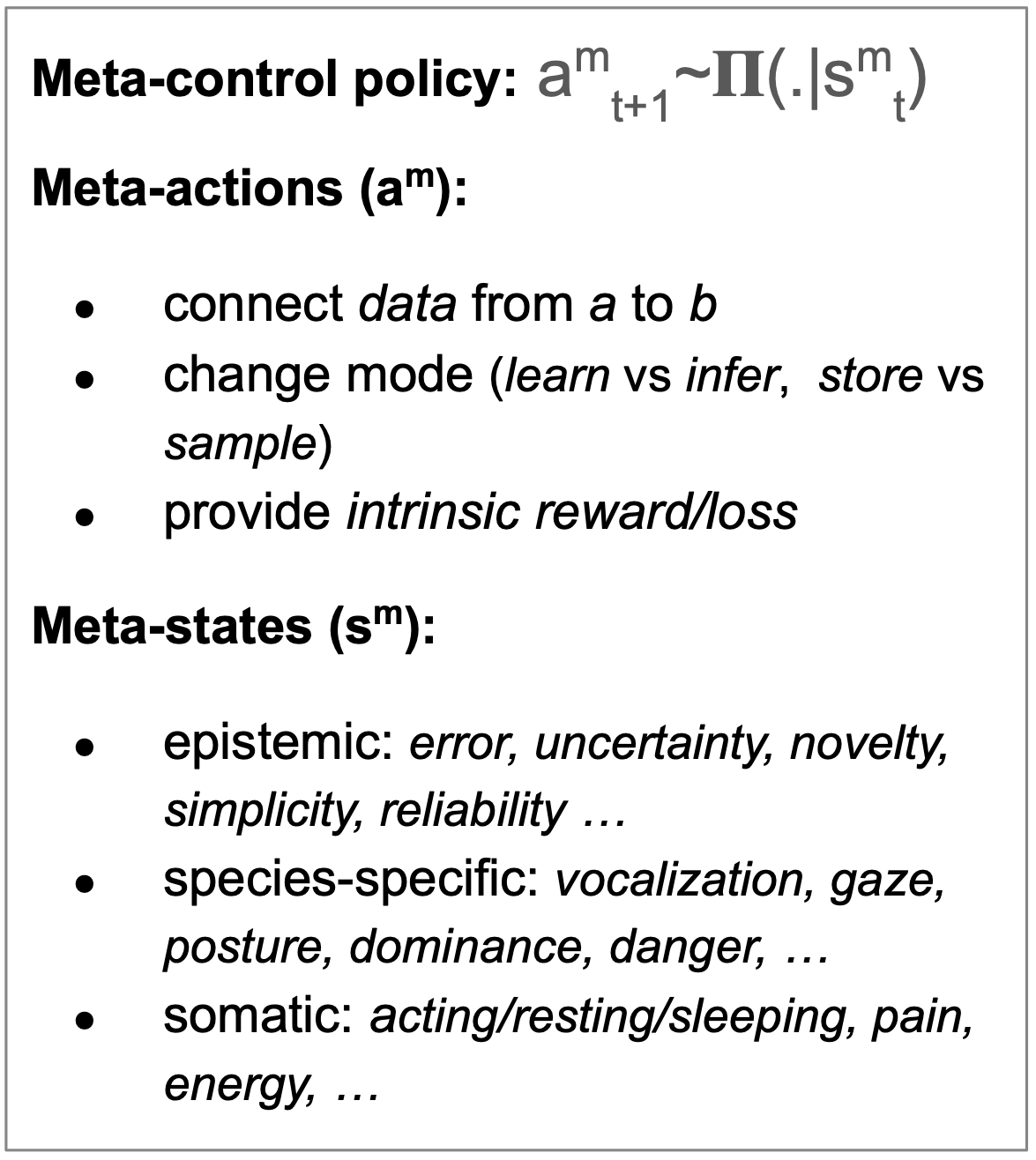}
    \includegraphics[width=0.6\linewidth]{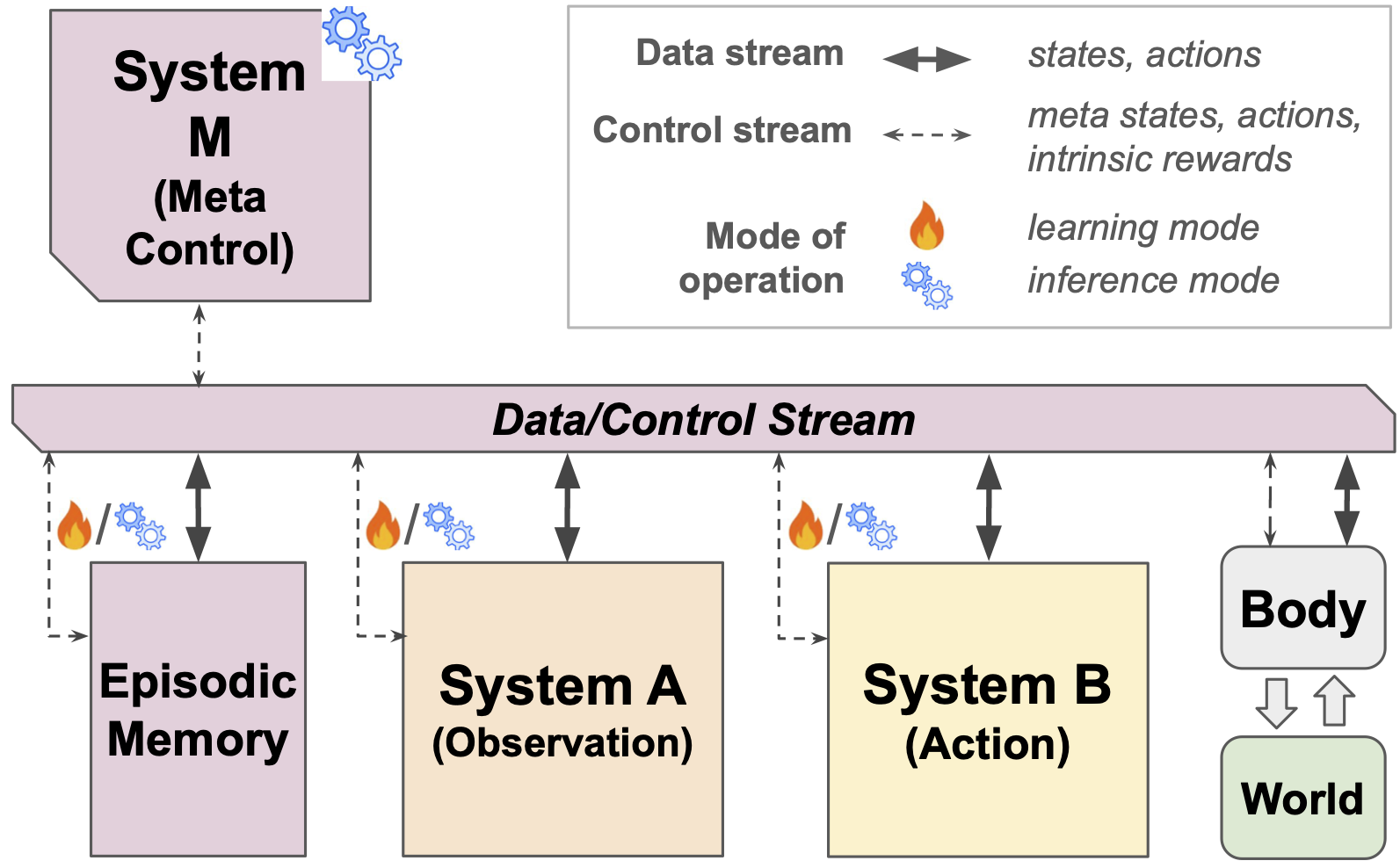}
    \caption{{\textbf{Blueprint of a cognitive architecture featuring System M as an autonomous orchestrator.} System M acts as a central control plane that automates data routing and training recipes. High-bandwidth \textit{data streams} (e.g., plain arrows) carry raw sensory inputs, motor commands, and latent representations between System A (perception/world modeling), System B (action/policy), and an episodic memory buffer. Low-bandwidth \textit{control streams} (e.g., thin dashed arrows) carry telemetry: System M monitors internal meta-states (such as prediction errors or uncertainty) and outputs routing commands (meta-actions) to dynamically open or close data pathways, effectively assembling and disassembling learning and inference pipelines on the fly. 
    }
    }\label{fig:systemM}
\end{figure}

A truly autonomous AI would integrate components that automate the traditional MLOps human functions of \textit{data sourcing and curation}, of building and adjusting \textit{training recipes}, and of \textit{benchmarking} performance and \textit{monitoring} learning signals \citep{Sculley2015Hidden,Paleyes2022Challenges}. In turn, this would require integrating in the learning architecture, an \textit{episodic memory} to store and replay raw or processed data, and a \textit{central orchestrator} to dynamically apply training recipes and route data streams.

At this stage, we can only offer a blueprint, a conceptual sketch requiring substantial future work to fully instantiate. 
We call the orchestrator \textbf{System M} (for Meta-control) (see Figure \ref{fig:systemM}). It would operate much like the Control Plane in \textit{Software-Defined Networking} \citep{Kreutz2015SDN}. It does not process the high-bandwidth \textit{data streams} of raw sensory inputs or motor commands directly. Instead, System M can be formalized as executing a (meta-)policy: it monitors low-dimensional telemetry or \textit{meta-states} (e.g., prediction errors, uncertainty, or somatic signals) and outputs \textit{meta-actions}. Paralleling the executive routing functions of the biological prefrontal cortex \citep{Miller2001PFC}, these meta-actions consist of dynamically interconnecting auxiliary processors (System A, System B) and episodic memory. By opening and closing data pathways, System M assembles and disassembles learning and inference pipelines on the fly.

Unlike the policies of System B, which are learned over the organism's lifetime via gradient descent or reinforcement, we propose that System M's core routing policy is hardwired---an evolutionarily fixed transition table that dictates when to explore, when to plan, and when to act\footnote{This may seem  a bit extreme considering that humans display the ability to acquire specific learning strategies through culture (eg, going to school). This is not incompatible with a fixed system M, however, as illustrated in Appendix \ref{app:advanced} where we show that advanced learning modes like learning by communication and by imagination can be implemented through a fixed system M.}. In the following sections, we first review how such meta-control functions in biological agents before mapping these principles to autonomous AI.

\subsection{Inspiration for Meta-Control in Humans and Animals}

Biological agents exhibit meta-control through multiple mechanisms that regulate learning and behavior in context-sensitive ways. These mechanisms can be grouped into three broad categories that parallel the engineering components of artificial systems: \emph{Input Selection}, \emph{Loss/Reward Modulation}, and \emph{Mode of Operation Control}. 

\textbf{Input Selection}. The bandwidth of sensory space is overwhelmingly large, requiring organisms to typically attend to only a subset of their data stream. For instance, infants preferentially attend to faces or vocalizations, a form of hardwired data curation that boosts System A's social and language learning \citep{morton1991,vouloumanos2007}.  Furthermore, they allocate attention to visual sequences of intermediate complexity, effectively generating a developmental curriculum that optimizes learning gains \citep{Kidd2012Goldilocks}. At a lower level, sensory streams are accompanied by internal uncertainty estimates, allowing the organism to selectively discount noisy modalities during multimodal integration \citep{Ernst2002Integration}. More generally, both children and adults actively explore and select their inputs during learning tasks, acting as intuitive scientists who sample data to resolve uncertainty \citep{Gureckis2012SelfDirected,gopnik2012scientific,Gopnik1999Scientist}.

\textbf{Loss/Reward Modulation}. The objective functions that organisms attempt to optimize are not fixed. At the developmental scale, critical periods illustrate that specific learning components are highly plastic only at certain developmental stages, effectively implementing a biological learning rate schedule \citep{Smith2005Embodied,Hensch2005}. Special modes of learning and memory consolidation are triggered during sleep or rest states \citep{Diekelmann2010Memory}. 
During activity, agents transition between exploratory behavior (sampling new options to reduce uncertainty) and exploitative behavior (maximizing known rewards) based on the volatility and predictability of their environment \citep{Daw2006ExplorationExploitation,Wilson2014ExploreExploit}. Social signals are also powerful modulators of learning strategies. Animals will learn preferentially from dominant conspecifics \citep{Kendal2015Dominance}. Children dynamically modulate their learning updates by prioritizing pedagogically cued demonstrations \citep{Csibra2009NaturalPedagogy}, but also learn according to estimated reliability and trustworthiness \citep{Harris2011SelectiveTrust}.

\textbf{Mode of Operation Control}. System A and System B can be run independently in inference or learning modes, with their inputs and outputs dynamically routed to each other or to episodic memory. Most spectacularly, during sleep, motor outputs and primary sensory inputs are actively gated off by the orchestrator, while episodic memory, System A, and System B remain highly active for both inference and offline learning, as evidenced by coordinated neural replay during REM and slow-wave sleep \citep{Rasch2013Sleep}.  During wake states, organisms flexibly arbitrate between \textit{habitual} control (System B reactive policies, or model-free behavior) and \textit{goal-directed} planning (System A world-model simulation, or model-based behavior) depending on reward stability, task volatility, and the degree of task mastery \citep{Daw2005Uncertainty,Dolan2013GoalHabit}. Other examples include flexibly toggling between observational social learning and direct trial-and-error problem solving depending on task novelty and success rates, as observed in corvids and macaques \citep{Subiaul2004MonkeysObservationalLearning,Taylor2012CrowsToolUse}. Similarly, human children leverage internal uncertainty estimates to switch into specialized learning modes, initiating exploratory play or engaging in metacognitive help-seeking when they recognize their own models are insufficient \citep{Lyons2010HelpSeeking}.

To summarize, meta-control can be modeled as a policy, $\pi(a^{m} | s^{m})$, which maps an internal \textit{meta-state} to a corresponding \textit{meta-action} (Figure \ref{fig:systemM}). The input meta-states can be summarized into three distinct types: \textit{epistemic signals}, which are derived by monitoring the internal operation of cognitive components (e.g., confidence, prediction error, learning gain, or novelty); \textit{species-specific signals}, which are high-priority environmental configurations recognized by pre-wired evolutionary detectors (e.g., direct gaze, dominance displays, looming stimuli, or heights); and finally, for embodied agents, \textit{somatic signals} derived directly from the physical body (e.g., energy levels, pain detectors, or arousal states). Meta-actions consist of dynamically connecting or disconnecting the input and output data streams of the subcomponents, turning them on and off in various operating modes (e.g., learning, inference, or optimization), providing them targets or internal rewards, and accessing episodic memory for memory replay or randomized batch learning. Together, these meta-actions enable the system to autonomously assemble and disassemble entire training and inference pipelines on the fly.

\subsection{Meta-Control in AI systems}

While a full specification of the possible routing circuits and meta signals is outside the scope of this paper, we refer the reader to Appendix \ref{app:advanced} for advanced operating modes, and Appendix C for further ethological examples of meta-signals. In addition, we provide below a non-exhaustive list of relevant work in the AI literature where isolated fragments of System M are actively being developed.

\textbf{Input Selection}. In machine learning, the autonomous curation of data streams is explored through the lens of \textit{Active Learning}, where epistemic meta-signals (like model uncertainty or ensemble variance) help query the most informative data points, drastically reducing sample complexity \citep{Ren2021DeepActiveLearning}.  In Reinforcement Learning, \textit{Prioritized Experience Replay} (PER), past experiences are sampled with a probability proportional to their temporal difference error (a direct equivalent of biological prediction error) \citep{Schaul2015PrioritizedReplay}. \textit{Mixtures of Experts} were an early implementation of dynamic routing of subsystems \citep{jacobs1991adaptive}, and have been applied to deal with noisy or incomplete modalities (e.g., \citealt{han2024fusemoe,mai2026umq}).

\textbf{Loss/Reward Modulation}. The autonomous, dynamic adjustment of objective functions is actively researched in the fields of \textit{Intrinsic Motivation} and \textit{Unsupervised Environment Design} (UED), where meta-states like novelty or prediction error, enable curiosity-driven exploration that mirrors biological exploratory play \citep{Oudeyer2007IntrinsicMotivation,Bellemare2016UnifyingCountBased,Pathak2017Curiosity}. \textit{Auto-Curriculum} algorithms dynamically modulate the difficulty of the training environment generating tasks at the frontier of the agent's capabilities \citep{Portelas2020AutomaticCurriculum}, while \textit{Continual Learning} methods like Elastic Weigh Consolidation mimick the biological critical periods discussed earlier \cite{Kirkpatrick2017EWC}.

\textbf{Mode of Operation Control}. The dynamic arbitration between fast inference (System B) and deliberate planning (System A) has initially been explored in RL through a meta-controller optimizing an 'imagination budget' by switching from world-model simulation (Monte Carlo Tree Search), to a reactive model-free policy based on task difficulty \citep{Hamrick2017Metacontrol}. Similarly, in Hierarchical Reinforcement Learning (HRL), a high-level manager policy does not output motor actions directly, but rather acts as a router that toggles lower-level, specialized sub-policies on and off depending on the task context \citep{Bacon2017OptionKeyboard}. These ideas are also explored through  \textit{inference-time compute scaling}. LLM-based reasoning and agentic models using \textit{'Thinking' Tokens} to solve complex tasks \citep{Yao2023TreeOfThoughts,Snell2024ScalingTestTime} can also be taught to switch between reactive responses and more expensive search \cite{Lin2023SwiftSage,Zelikman2024QuietSTaR}.

Despite these advances, modern AI still lacks a unified Control Plane that integrates all these meta-functions—input selection, reward modulation, and operational routing—into a single, cohesive architecture.
Next, we explore how such a system could be built.

\section{Bootstrapping Autonomous Learning: An Evolutionary-Developmental Framework}\label{sec:evodevo}


Our proposed architecture with Systems A, B, and M is an overall blueprint for autonomous learning. Building a functional model may prove challenging due to the interdependencies between the three components:
if System A relies on action-generated data to acquire grounded representations, and System B in turn depends on perceptual structure to guide efficient action, how can either system be initialized so that learning can begin?
Likewise, if System M is crucial for orchestrating learning in the other systems, but itself depends on well-calibrated uncertainty or error signals produced by them, how can it be set up to ensure robust learning across diverse environments?
In the following sections, we draw from biology the distinction between adaptations at the evolutionary versus developmental scale (Evo/Devo), to outline a strategy for resolving this chicken-and-egg (and rooster) problem.

\subsection{Evo/Devo Scales for Organisms}

Even a cursory examination of natural organisms reveals that none start from randomly initialized neural networks \citep{Zador2019}. Animals inherit a highly specified species-typical nervous system that unfolds over developmental time. This inherited structure constrains and guides learning by providing \textit{inductive biases} that shape what can be learned, how rapidly, and through which modalities \citep{KarmiloffSmith1992,Johnson2001,Gallistel1990,Gallistel2013}. In computational terms, such biases can be understood as an initial \textit{architectural} and \textit{parametric configuration}, coupled with a developmental program that provides an \textit{internal curriculum} \citep{Bengio2009CurriculumLearning} that gradually increases the complexity of perception, action, and learning \citep{ThelenSmith1994}. Empirically documented mechanisms include synaptic growth and pruning, temporally regulated plasticity, critical periods, spontaneous neural activity \citep{Huttenlocher1979,Hensch2005,HaddersAlgra2018,molnar2020transient}, and  progressive increase in visual acuity and motor degrees of freedom \citep{TurkewitzKenny1982,turvey1990coordination}. Through these processes, organisms acquire increasingly sophisticated representations and action policies starting from a deliberately simplified regime \citep{Elman1993}, but by no means a tabula rasa state.

One could argue, however, that this merely shifts the bootstrapping problem from ontogeny to phylogeny, without explaining how inductive biases and developmental curricula arise in the first place. Here we can offer only speculative remarks. Large-brained, highly behaviorally flexible organisms are evolutionarily recent, and early life forms likely operated with comparatively simple sensorimotor loops in environments where modest behavioral plasticity was sufficient \citep{striedter2005principles,paulin2021events,keijzer2015moving}. 
Evidently, large brains are not a systematic outcome of evolution, as brainless organisms still occupy the majority of the biomass \citep{bar2018biomass}. Nevertheless, evidence suggests that during the Cambrian, a fraction of this biomass transitioned into more cognitively sophisticated organisms with improved sensory and effector organs  (see a review in \citealt{hsieh2022phanerozoic}) correlatively to increasingly complex ecosystems including predation and competition \citep{plotnick2010information}. The putative relationship between niche complexity, brain complexity and learning abilities remains a fascinating and controversial topic \citep{krause2022evolution,dukas2019cognitive,arendt2016nerve}, but suggests a pathway from simple to increasignly sophisticated autonomous learning capabilities.

\subsection{Evo/Devo for autonomous AI}

In machine learning, the evolutionary–developmental distinction is also present, but instantiated in strikingly different terms. The developmental scale is the realm of current machine learning algorithms, as discussed above as System~A and B. The evolutionary scale is, for the most part, manifested through the standard practices of scientific research (distributed research teams, publications, open sourcing, etc.). In this sense, System~M is implemented through humans—research scientists, engineers, and students. 

\begin{figure}[t]
    \centering
    \includegraphics[width=15cm]{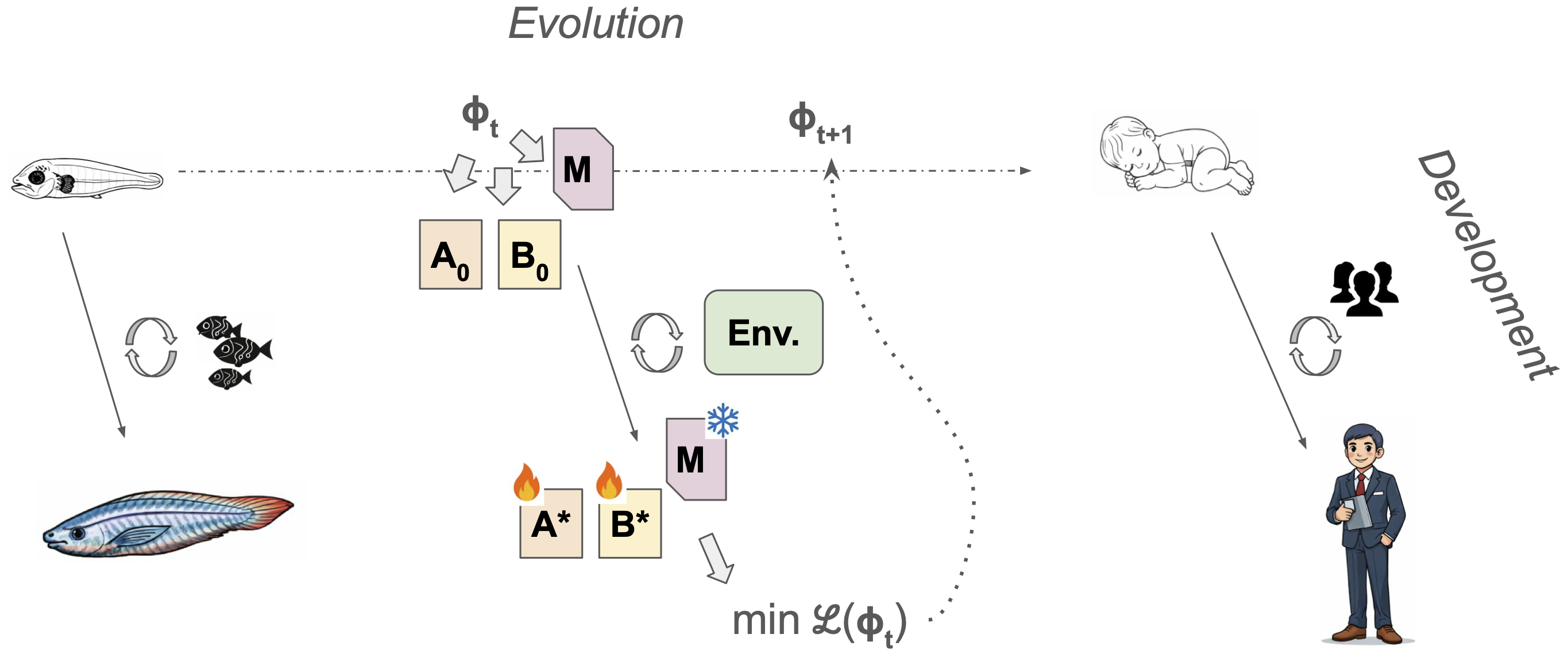}
    \caption{Evo/Devo framework for building autonomous learning agents. Learning takes place at two scales. In the developmental scale, the learner's architecture (A, B and M) is initialized from meta parameter $\phi$. A and B update their parameters through interaction with the environment controlled by a fixed controler M. In the the evolutionary scale, $phi$ is updated to optimize a fitness function $\mathcal{L}$ measured over the life cycle of the system. \textit{(images from ChatGPT)}. }\label{fig:evodevo}
\end{figure}

Here we propose to formalize the problem in a unified fashion
(Figure \ref{fig:evodevo}): each agent is defined by a set of parameters $\phi$ that correspond to the information in its genetic code. At "birth", $\phi$ is used to specify an architecture comprising Systems A, B, and M with their initial parameters ($A_0$,$B_0$,$M_0$). During the developmental scale (inner loop), the agent learns in interaction with its environment by updating Systems~A and B as controlled by a fixed System~M. On the evolutionary timescale (outer loop), $\phi$ is optimized through an externally handcrafted fitness function $\mathcal{L}$ computed at the end of the life cycle of each agent. The fitness function (and the environments) are the only handcrafted parts of the system. The update rules, initialization, data filtering, curriculum, etc., are all provided through System~M.

\begin{table}[h!]
\center
\begin{tabular}{lp{10cm}}
&$\phi_{t+1}=\arg\min_{\phi_{t}} \mathcal{L}(A_0:A_K,B_0:B_K)$\\
subject to \\
&$A_0,B_0,M=\operatorname{Init}(\phi_t)$\\
&$A_{i+1},B_{i+1}=\operatorname{Update}(M,A_i,B_i,\mathit{Env})$\\
where\\
&$\operatorname{Init}$ is the initialization procedure\\
&$\operatorname{Update}$ is the inner loop update rule\\
&$\mathit{Env}$ is the interactive environment\\
\end{tabular}

\end{table}

This problem belongs to the class of \textit{bilevel optimization} problems. Methods to solve it include research on meta-learning \citep{Schmidhuber1987EvolutionNN,Thrun1998LearningToLearn,Andrychowicz2016LearningToLearn,Finn2017MAML}, neural architecture search \citep{Stanley2002NEAT,Zoph2017NAS}, curriculum learning \citep{Bengio2009CurriculumLearning,Graves2017AutomatedCurricula}, intrinsic motivation \citep{Oudeyer2007IntrinsicMotivation}, exploration strategies \citep{Bellemare2016UnifyingCountBased,Pathak2017Curiosity}, and continual learning \citep{ring1997child,thrun1998lifelong,de2021continual}, although these elements have not yet been integrated to deliver an autonomous learning system.

Note that bilevel optimization problems are challenging and have typically been studied in relatively simple situations. One challenge is that at the outer level, a whole life cycle is just one data point. In order to optimize $\phi$, one needs to run millions of simulated life cycles which themselves imply learning over millions of datapoints. This requires considerable feats of improving memory and compute efficiency of the basic architecture. In addition, while bilevel optimization can be solved using a variety of techniques, both gradient-based and gradient-free methods (see \citealt{Sinha2018} for a review), extending such techniques to much larger architectures yields severe scalability issues (see \citealt{Lorraine2020, Real2019, Metz2021}). A second challenge is that what has to be optimized is itself a dynamic system comprising a learner and an environment in interaction. Here, we would suggest that an \textit{Evolutionary Curriculum}, gradually increasing the diversity and unpredictability of the environments, would help allowing the three components to co-evolve and solve the chicken–egg–rooster problem \citep{Oudeyer2007IntrinsicMotivation,Leike2017AIComplete}.

Before closing, let us note that we are not proposing that the full complexity of living organisms need to be reproduced to construct autonomous systems. These principles can be applied to systems behaving adaptively in simpler environments as has been done with adaptive control, for instance. One can view our proposal as an extension of adaptive control to more complex problems with the tools of modern AI.

\section{Conclusions}\label{sec:conc}
Contrary to animals, current AI systems don't learn autonomously. Their learning is restricted to an off-line MLOps pipeline involving a large team of experts who prepare the data, build the training recipes and adjust them according to performance metrics. Once deployed, the models do not learn anymore, and adaptation to specific use cases is on the users through prompting or fine tuning. 

Our analysis shows that current AI systems are lacking are three key abilities that are found across the animal kingdom: the ability to select their own training data (\textit{active learning}), the ability to flexibly switch between learning modes (\textit{meta control}), the ability sense their own performance (\textit{meta-cognition}). In this paper, we identified three roadblocks that stand in the way to unlocking these missing abilities: the necessity to merge AI techniques from well siloed paradigms (self-supervised learning, reinforcement learning), the necessity to build an integrated cognitive architecture that automates the MLops pipeline through additional data routing, orchestration, and internal sensing components (the A-B-M architecture), the necessity to build these components jointly through an evolutionary/development scheme using simulated environments. 

\subsection{Why is Autonomous Learning Useful?}

Systems that learn autonomously and adaptively like animals or humans could offer a powerful leap toward more general, robust, and flexible intelligence. Such systems would learn from raw experience, side stepping the entire MLOps pipeline, enabling them to operate in complex, changing, or poorly understood environments. Like children or animals, they could generalize from limited examples, explore new tasks through curiosity, and develop an embodied, grounded understanding of the physical and social world \citep{Lake2017BuildingMachinesLikePeople}. These features would make autonomous AI well-suited for real-world deployment in uncertain or dynamic settings—ranging from home robots to scientific discovery.

Beyond practical utility, they also offer a unique opportunity to reverse-engineer natural intelligence, accelerating the scientific understanding of
the neural mechanisms and developmental trajectories of cognition \citep{Turing1950ComputingMachinery,Hassabis2017NeuroscienceInspired} across different environments and cultures. 

\subsection{Why is it Difficult?}

Building a machine that learns like children do has been envisioned since the inception of AI \citep{Turing1950ComputingMachinery}, but several technical and ethical challenges remain.

\textbf{Simulators and Environments}. Training tightly coupled Systems A, B, and M requires environments that should both be realistic and optimized for speed. On the realism side, it should provide diverse sensory modalities to enable observational learning (System A), rich embodied affordances and goal-conditioned tasks to support interaction and control (System B), and sufficiently diverse and non-stationary dynamics to train a robust meta-controller (System M). On the speed side, faster than real time is necessary for the evolutionary scale to be applicable. Procedural generation can help scale diversity, but requires careful design to avoid trivial patterns or degenerate exploration \citep{Cobbe2020Procgen}. Incorporating social agents or enabling teacher–learner interactions is especially challenging at scale \citep{Crosby2019AnimalAI,Savva2019Habitat}.

\textbf{Evaluation and Benchmarking.} As agents become more general, task-specific benchmarks lose their diagnostic value, creating the need for new evaluation paradigms. We propose distinguishing between \textit{unit tests}, which assess individual learning components in isolation, and \textit{integration tests}, which evaluate their combined behavior.

Unit tests should target the core competencies of each system: sample-efficient perceptual generalization for System A; few-shot task adaptation under sparse rewards for System B; and, for System M, efficient switching between learning modes in non-stationary environments, as well as the emergence of more advanced capabilities such as learning through communication or imagination.

Integration tests should assess end-to-end learning performance in realistic settings. One promising approach is to compare humans and AI agents in terms of learning speed on novel tasks—for example, the number of trials required to learn a new videogame \citep{Lake2017BuildingMachinesLikePeople}, or the number of hours of exposure needed to acquire a language at a level comparable to that of human children \citep{Dupoux2018}. Such benchmarks should emphasize few-shot, data-efficient generalization and closely mirror developmental learning processes \citep{Hill2020GroundedLanguage,ChevalierBoisvert2019BabyAI}.

\textbf{Scaling Bi-Level Optimization.} Optimizing lifelong learning processes in complex environments is both computationally demanding and highly sensitive to curriculum design. Addressing these challenges requires progress along several complementary directions. First, we need more efficient inner-loop learners with low data and computational requirements. Second, meta-objectives must be designed to effectively shape priors, intrinsic reward structures, and communicative tendencies. Third, appropriate strategies for curriculum scheduling and environment sampling are necessary to accelerate the emergence of robust, autonomous learning architectures.

\textbf{Ethical Issues.} Developing AI systems that learn in ways analogous to humans or animals raises novel ethical concerns that go beyond those associated with current AI technologies. In particular, autonomous learning introduces new trade-offs between flexibility, safety, and societal oversight.

A first challenge concerns the tension between \textit{adaptability} and \textit{controllability}. As systems are granted greater autonomy in exploratory learning modes, it becomes harder to guarantee that they remain aligned with intended objectives. Mitigating this risk may require explicit auditing mechanisms and the ability to intervene in or constrain the meta-control system (System M).

A second risk is \textit{alignment hacking}. Although animals evolved to optimize reproductive fitness, their everyday behavior is often driven by proxy objectives such as exploration or play, and can occasionally give rise to maladaptive outcomes, including addiction or self-harm. These behaviors arise because biological agents optimize internally generated signals that may become mismatched to their environment \citep{Tooby1992EnvironmentOfEvolutionaryAdaptedness,Buss2019EvolutionaryPsychology}. Autonomous artificial agents that rely on similar proxy signals may face analogous vulnerabilities.

A third concern relates to \textit{over-trusting}. As artificial agents become more human-like in their behavior and learning trajectories, users may increasingly anthropomorphize them, leading to emotional attachment, misplaced trust, or opportunities for manipulation \citep{Turkle2011AloneTogether}. Addressing this risk requires transparency about system capabilities and limitations, as well as mechanisms that ensure meaningful societal and user control.

Finally, autonomous learning systems often depend on bodily or somatic signals to guide adaptation. To the extent that these signals are processed in ways functionally analogous to pain or fear in biological organisms, this raises unresolved questions about the \textit{moral status} of such agents \citep{Gunkel2012MachineQuestion,Birhane2021EthicsEmbodiedAI}.

\section{The Path Forward}

To build agents capable of autonomous, open-ended learning, we must move beyond disjointed, hand-designed
training paradigms and rigid execution. Already, the AI field is moving beyond fixed systems with frontier topics such as \textit{runtime adaptation} and \textit{inference-time compute}, including Large Language Models (e.g., test-time training with verifier-driven selection \citep{moradi2025vdsttt}, synergistic adaptation \citep{xu2025sytta}, and adaptive retrieval \citep{sun2026ttarag}), Vision-Language Models \citep{lei2025metatpt,kojima2025lorattt}, speech recognition \citep{fang2026asrtra}, and core Reinforcement Learning \citep{chehade2025tram,bagatella2025goalrelated}.
Closely related ideas are also gaining traction in robotics under adjacent terminology such as online adaptive learning \citep{yuan2026adaworldpolicy}, deployment-time correction \citep{welte2026flowcorrect}, test-time reinforcement learning \citep{liu2026ttvla}, and test-time mixture of world models \citep{jang2026tmow}. 
These adaptations, however exciting, are still minor variations over an overall rigid system, compared to children who learn a whole language and new skills at ’test time’.

The challenges are considerable and we are probably decades away from fully autonomous, broad scope learning systems.  Our proposed architecture (System A-B-M) is a tentative blueprint that we hope can inspire research by providing a unified conceptualization of this problem space, and a path towards building actual autonomous learning systems---inspired by natural intelligence, but not strictly bound to replicate it. The alignment of such adaptive systems with humans goals, and the autonomy--controllability trade-offs, are of paramount importance, and should be considered within an evolutionary--developmental framework through the careful design of fitness rewards and interactive environments.
But even before fully autonomous learning systems are achievable, the successes and failures in building such systems will be scientifically invaluable, providing quantitative models of how biological organisms successfully learn and adapt in the wild, and offering insights on the very nature of learning and intelligence.

\newpage
\appendix
\section*{Appendix}
\renewcommand{\thesubsection}{\Alph{subsection}}
\counterwithin{figure}{subsection}
\counterwithin{table}{subsection}


\subsection{What's the link with System 1 and 2?}\label{sec:system12}

Kahneman's \citep{Kahneman2011Thinking} distinction is about modes of inference (fast parallel inference, as in a forward pass in a DNN versus slow multistep inference, as in tree search or chain of thought). Our distinction is about modes of learning (from static data versus from interaction). Even though these distinctions seem similar, one could argue that they are actually orthogonal: one could learn from static data through simple backpropagation, or after having run simulation or imagination steps; conversely, even though learning from interaction does require at least one step of action, this action step could be generated reflexively through a learned policy rather than search.

\begin{table}[h]
\begin{tabular}{llll}
\hline
&& \multicolumn{2}{c}{\textit{Modes of learning}}   \\
\cline{3-4}
&         & \multicolumn{1}{c}{System A}                                  & \multicolumn{1}{c}{System B}                  \\
\hline
\multirow{2}{*}{{\begin{tabular}[c]{@{}l@{}}\textit{Modes of }\\ \textit{inference}\end{tabular}}}&System 1 & predictive coding; statistical learning   & policy learning           \\
&System 2 & counterfactual reasoning; causal learning & learning through planning\\
\hline
\end{tabular}
\caption{Link between System 1/2 and System A/B}
\end{table}

\subsection{Two Advanced Meta-Controlled Learning Modes: Communication and Imagination}\label{app:advanced}

In species living in complex and unpredictable environments, a capable System M enables new functional configurations, which we illustrate briefly here.

\subsubsection{Learning from Communication}

One of the functions of System M is to select “important” inputs to learn from, and across many species, social context is highly salient. We call this learning through communication, and it serves as a mechanism for accelerated, distributed learning \citep{Tomasello1999CulturalLearning,Heyes2018CognitiveGadgets}.

This ranges from basic attention-based mechanisms to complex cultural transmission. Below is a structured summary of key forms of social learning, grouped by increasing complexity.

\textbf{Basic Observational and Associative Learning.} Individuals adjust their behavior by noticing where others direct their attention or how they react to stimuli. Example: A child becomes interested in a toy only after seeing another child play with it \citep{Bandura1977SocialLearning,Moore2013SocialLearningReview}.

\textbf{Behavioral Copying.} Learners replicate others’ actions or outcomes, either by imitating specific movements or by finding new ways to reach the same goal. Example: A chimpanzee uses a different method to retrieve food after seeing another (typically dominant) chimpanzee succeed \citep{Whiten2005ChimpCulture,Subiaul2004MonkeysObservationalLearning}.

\textbf{Guided Learning.} Learning is supported by intentional or structured social interaction with a teacher, using methods such as shared attention, social referencing, behavioral scaffolding, or demonstrations. Example: A parent shows how to use an object while making eye contact with a child \citep{Bruner1983ChildsTalk,Wood1976Scaffolding}. Such a mode of learning can extend to mimicking gestures that have no obvious goals (like learning rituals or complex recipes).

\textbf{Higher-Level Social Learning.} Learners internalize norms, generalize across contexts, and pass on knowledge culturally through abstract symbolic formats like language \citep{Csibra2009NaturalPedagogy,Henrich2016SecretOfSuccess}. Example: Children are explicitly taught about social rules, follow verbal instructions.

While the first two forms of social learning have been documented across many species (mammals and birds), guided learning is more rarely observed, and only humans exhibit the higher-level form of social learning \citep{Tomasello2009WhyWeCooperate}.

System M supports learning through communication by attending to communicative triggers (e.g., pointing, direct gaze, imperative intonation), and routing the highlighted inputs for System A or B learning. The strength of this learning episode can be one-shot and modulated by System M based on perceived social importance or trust in the teacher (epistemic vigilance; \citealp{Sperber2020EpidemicOfRumours}).

In standard AI systems, learning through communication is done externally, through a team of data scientists curating the data from reliable sources, or through post-training methods. But because this process is entirely externalized, current systems are unable to learn socially or exert epistemic vigilance regarding the source of their data.

\subsubsection{Learning from Imagination}

Another mode of operation enabled by System M is the ability to learn from internally generated inputs (learning from imagination). This mode of operation has been documented in varying degrees of complexity across species.

\textbf{Memory replay at rest.} During pauses in activity (e.g., at decision points or after receiving a reward), rodents experience a reactivation of the place cells they have recently visited, in the same order or reverse order, typically at a much faster rate. This has been linked to decision making, near-term planning, and updating value functions in reinforcement learning \citep{Foster2006ReverseReplay,Diba2007ForwardBackwardReplay,Mattar2018PlanningReplay}.

\textbf{Memory replay during sleep.} During non-REM sleep, recent experiences are replayed at a compressed time scale, often in longer episodes including novel combinations. This has been linked to memory consolidation and the formation of long-term schemas or generalization \citep{Wilson1994HippocampalTheta,Diekelmann2010Memory,Kumaran2016MemoryReplay}.

\textbf{Long-horizon planning.} Humans and some animals show evidence of problem-solving without prior trial and error in tasks involving multiple steps of tool use, delayed gratification, or counterfactual reasoning \citep{Suddendorf2007EpisodicFuture,Redshaw2016FutureThinking,Tulving2005EpisodicFutureMemory}. Such phenomena highlight imagination-like simulation as a substrate for flexible foresight

System M can support these modes of operation by switching Systems A and B into inference mode, routing input information from memory (instead of the sensors), and routing output information (e.g., actions) to internal simulation. It can then trigger learning on the successful imagined trajectories. This highlights the flexibility of System M—not just as a router, but as an enabler of qualitatively new learning regimes \citep{Ha2018WorldModels,Hafner2020dreamer}.

\subsection{Examples of Meta-Control signals for autonomous learning in Humans and Animals}


In Table \ref{tab:animalM}, we list examples of meta-states that have been found in humans and animals to have direct effect on input selection (attention, gaze), on learning efficiency (loss/reward modulation) or on mode of Control. We sort them according to our three way category of species-specific, epistemic and somatic signals. Interestingly, most of these signals are relatively simple to compute or rest on rudimentary computations (for instance, the preference for faces boils down to a preference to for a dark T pattern on a white background), making it relatively simple to emerge in a meta learning setting. 
They could inspire the design of meta-controlers in artificial agents depending on the application. 
\begin{table}[]
\begin{tabular}{p{0.1cm}p{3.2cm}p{2.5cm}p{8.9cm}}
\hline
    & \textbf{Meta-states} & \textbf{Species}  &\textbf{Effects} (sample references) \\
\hline
\multicolumn{2}{l}{\textbf{Species-Specific signals}} \\
    & Faces and\newline vocalizations           & Human infants                &IS \citep{Johnson1991Faces,vouloumanos2007};  LE \citep{ferry2010words}  \\
    & Gaze direction       & Human infants, dogs, corvids &IS (gaze following: \citealt{farroni2002eye,tomasello1998five}) \\
    & Pedagogical signals\newline \textit{(high pitch, direct gaze; pointing)}    & Human infants & IS  \citep{fernald1985four}, LE \citep{thiessen2005infant}, MC (shifts from memorization to generalization; \citealt{Csibra2009NaturalPedagogy,Gergely2002RationalImitation})  \\
    & Self-propelled,\newline biomechanical\newline motion  & Human infants, chicks    &IS  \citep{simion2008predispositions,scholl2000perceptual}, LE  \citep{nairne2013animacy}; MC (switches to teleological reasoning; \citealt{csibra1999teleological}) \\
    & Potential threats\newline \textit{(looming, snake or\newline spider-like objects)}  & Human infants, mammals & IS \citep{ball1971infant,yilmaz2013looming,deloache2009detecting}; LE  \citep{ohman2001fears,cook1989observational}; MC (freezing/high vigilance;  \citealt{fanselow1994behavior})\\
    & Dominant,\newline prestigious or in-\newline group conspecifics  & Human infants, primates & IS \citep{shepherd2006social,chudek2012prestige,kinzler2007native}; LE \citep{buttelmann2013fourteen}; MC (shifts from high to low level imitation; \citealt{haun2014children}) \\
\hline
\multicolumn{2}{l}{\textbf{Epistemic signals}} \\
    & Reliable conspecifics / Selective trust           & Human infants, primates, dogs & IS \citep{poulin2011infants,schmid2017great}; LE \citep{koenig2004trust,zmyj2010fourteen}; MC (shifts to exploration when unreliable; \citealt{gweon2014sins,takaoka2015do})\\
    & Logical/semantic contradictions or conflicts & Human adults, infants, primates & IS \citep{baillargeon1985object}; LE??  \citep{stahl2015observing}; MC (switches to world-model simulation; \citealt{botvinick2001conflict}) \\
    & Unexpected outcomes & Mammals, birds, insects & IS \citep{sokolov1963higher}; LE \citep{pearce1980model,schultz1997neural}; MC (shifts to exploration: \citealt{dayan2002reward}) \\
    &Uncertainty & Human infants, primates & LE (bayesian multimodal fusion; \citealt{Ernst2002Integration}); MC (halts exploitation (opt-out); \citealt{goupil2016infants,hampton2001rhesus}) \\
            & Stimuli of intermediate complexity  &Human infants, primates  & IS (\citealt{Kidd2012Goldilocks}), LE (\citealt{kang2009the})\\
\hline
\multicolumn{2}{l}{\textbf{Somatic signals}}\\
            & Sleep & All animals &MC, LE, IS  (disconnection of sensory inputs, boost in learning, memory replay) \citep{buzsaki2015hippocampal,tononi2014sleep}\\
            & Rest  & All animals  & MC, LE, IS  (sensory input attenuation, learning by imagination, memory replay) \citep{kam2011slow,buckner2008brains,raichle2015default}\\
            & Pain & All animals &  MC (interrupts all goal-directed plans; switch to reactive actions, followed by replay and world modeling), LE (single shot learning) \citep{eccleston1999pain} \\
            & Hunger & All animals & MC (high exploration, boost in goal directed), IS(enhanced detection of food) \citep{balleine1998goal,kolling2012foraging}. \\
            & Stress & All animals & MC (low stress: world modeling simulation, exploration; high stress: reactive policies, exploitation) \citep{schwabe2009stress} \\

\hline
\end{tabular}
\caption{Non exhaustive list of meta-states in humans and animals, and their effects interpreted in terms of Input Selection (IS: attracting gaze or attention), Learning Efficacy (LE: boosting learning rate) and Mode Control (MC) in humans and animals.}\label{tab:animalM}
\end{table}

\section*{Acknowledgements} 
This paper is the result of several years of discussion between the authors and an interdisciplinary workshop on autonomous learning at META in July 2, 2025 in New York. We thank the participants of this workshop, in particular Karen Adolf, Catherine Tamis Lemonda, Pulkit Agrawal, Carl Vondrick, Linda Smith, Allison Gopnick, Brenden Lake, Mido Asran, Michael Henaf, Alessandro Lazaric, Mahi Luthra, Juan Pino, Jiayi Shen and Pascale Fung who shared useful insights. We are especially grateful to Shiry Ginosar for her detailed, incisive, and highly useful comments on an earlier version of this paper. 
A first version of Section 2 has been published as a section in  \cite{fung2025embodied}. The work was conducted while JM and YLC were at META. ED
in his EHESS role was supported by the Agence
Nationale pour la Recherche (ANR-17-EURE0017 Frontcog, ANR10-IDEX-0001-02 PSL*) and an ERC grant (InfantSimulator), these granting agencies declining responsibilities for the views and opinions expressed. AI tools were used in the conception stage, to discuss the logic and structure of the paper, to produce the graphical elements of figures 1 and 3, and in the final stage for a stylistic / grammatical check and rechecking of the bibliography.

\newpage
\bibliographystyle{plainnat} 
\bibliography{paper.bib}

\end{document}